\documentclass{article}
\usepackage{ragged2e}
\usepackage{hyperref} 
\usepackage{orcidlink} 
\usepackage{spverbatim}
\usepackage{changepage}
\usepackage{algorithmic}
\usepackage{tabularx}
\usepackage{academicons}
\usepackage{array}
\usepackage{multirow} 
\usepackage{booktabs}
\usepackage{tikz}
\usepackage{pgfplots}
\pgfplotsset{compat=1.18} 
\usepackage{algorithm}

\title{Intelligent Video Recording Optimization using Activity Detection for Surveillance Systems}

\begin{document}

\maketitle

\centering

\author{
Youssef Elmir\\ 
    \thanks{Laboratoire LITAN\\ École supérieure en Sciences et Technologies de l’Informatique et du Numérique, RN 75, Amizour, 06300, Béjaia, Algérie. 
    \href{mailto:elmir@estin.dz}{elmir@estin.dz}.} \\
    \and 
    Hayet Touati\\
    \thanks{École supérieure en Sciences et Technologies de l’Informatique et du Numérique, RN 75, Amizour, 06300, Béjaia, Algérie. \href{mailto:h_touati@estin.dz}{h\_touati@estin.dz}}
    \and 
    Ouassila Melizou\\
    \thanks{École supérieure en Sciences et Technologies de l’Informatique et du Numérique, RN 75, Amizour, 06300, Béjaia, Algérie. \href{mailto:h_touati@estin.dz}{o\_melizou@estin.dz}}

}
\raggedright
\justifying 

\begin{abstract}
Surveillance systems often struggle with managing vast amounts of footage, much of which is irrelevant, leading to inefficient storage and challenges in event retrieval. This paper addresses these issues by proposing an optimized video recording solution focused on activity detection. The proposed approach utilizes a hybrid method that combines motion detection via frame subtraction with object detection using YOLOv9. This strategy specifically targets the recording of scenes involving human or car activity, thereby reducing unnecessary footage and optimizing storage usage. The developed model demonstrates superior performance, achieving precision metrics of 0.855 for car detection and 0.884 for person detection, and reducing the storage requirements by two-thirds compared to traditional surveillance systems that rely solely on motion detection. This significant reduction in storage highlights the effectiveness of the proposed approach in enhancing surveillance system efficiency. Nonetheless, some limitations persist, particularly the occurrence of false positives and false negatives in adverse weather conditions, such as strong winds.

\end{abstract}


\section{Introduction}
The widespread installation of surveillance cameras has led to a significant increase in visual data, with estimates predicting over 1.4 billion cameras worldwide by 2024, generating vast amounts of video data daily. This surge poses major challenges in terms of storage, requiring hundreds of petabytes to manage this critical information. Efficient methods to manage and facilitate the search of this data have become imperative.

Surveillance cameras, essential for security, create an information overload. A single HD camera can produce nearly 650 megabytes of data per minute, resulting in a substantial data management challenge when multiplied across thousands of cameras in urban areas. Much of this data is redundant or irrelevant, necessitating careful consideration of optimal storage methods to ensure the rapid retrieval of important information.

The key problem is optimizing the storage of surveillance camera data to prevent storage congestion while maintaining necessary recordings. Object and motion detection algorithms emerge as promising solutions, filtering sequences to record only significant activities. This approach addresses the challenge of information overload in video surveillance.

This study aims to optimize storage space through innovative methods using motion and object detection algorithms and to implement a solution capable of discriminating important scenes. This dual objective seeks to balance storage efficiency with the relevance of the recordings. The structure of the remain of this paper is as follows: Section 2 reviews related works in intelligent surveillance systems, comparing their strengths and weaknesses. In section 3, a proposed methodology of activity detection is introduced with a overall architecture of the proposed system. Implementation is presented in section 4, and experiment results and discussion in section 5. The paper concludes with a summary of findings and a discussion of future research directions.

\section{Related Works}
Previous studies have explored various methods for optimizing video recording. Background subtraction and frame differencing are commonly used for motion detection, while object detection methods like Faster R-CNN and YOLO have shown promising results in identifying specific objects in video streams.

\begin{itemize}

\item Arham et al \cite{arham2023motion} developed a comprehensive real-time object detection system integrating motion detection, face detection, and human activity recognition, effective in real-world applications but lacking comparison with existing systems and detailed dataset descriptions.

\item Pal et al \cite{pal2023object} proposed a composite block matching algorithm for efficient motion estimation in video sequences, enhancing accuracy and processing speed but introducing computational complexity that may limit real-time applicability.

\item Sadoun et al \cite{Abdelbaki2016} addressed challenges such as illumination changes and shadow detection, creating a background modeling and subtraction algorithm. The model detects mobile objects but struggles with scene changes and fixed cameras, needing more quantitative measures.

\item Sreenu et al \cite{sreenu2019intelligent} reviewed deep learning techniques in intelligent video surveillance, highlighting advancements and challenges like computational complexity and the need for large datasets. They called for more robust models and multi-sensor data integration.

\item Xia et al \cite{9043535} presented a hybrid LSTM-CNN model for human activity recognition, showing superior accuracy but facing high computational costs and the need for extensive labeled data.

\item Adarsh et al \cite{adarsh2020suspicious} proposed a model using YOLO and ResNet-34 for detecting suspicious behavior, achieving significant precision but requiring substantial computational resources.

\item Lys et al \cite{lys2023development} developed a motion detection and object recognition system using OpenCV but lacked detailed results on detection efficiency.

\item Alajrami et al \cite{Alajrami2019OnUA} proposed AI techniques to enhance human identification in surveillance, improving accuracy and speed but facing high computational demands and privacy concerns.

\item Ullah et al \cite{ullah2019activity} combined CNN and LSTM for human activity recognition, improving accuracy but facing computational challenges and the need for broader dataset evaluation.

\item Boumediene et al \cite{boumediene2022detection} created a real-time object detection model using YOLO, achieving good accuracy but requiring a more diverse dataset for better generalization.

\item Suradkar et al \cite{suradkar2015automatic} proposed an automatic motion detection system improving surveillance efficiency but needing evaluation in varied environments.

\item Dhulekar et al \cite{dhulekar2017motion} developed a two-step system for motion and object detection, suggesting deeper learning techniques for more complex object detection.

\item Kapania et al \cite{kapania2020multi} combined YOLOv3 and RetinaNet for robust object detection and tracking, demonstrating high performance but suggesting YOLOv3 for speed and efficiency.

\item Dave et al \cite{dave2022gabriellav2} proposed a real-time action detection system addressing class imbalance and multi-label actions, achieving state-of-the-art performance but needing more representative datasets.

\item Bosquet et al \cite{bosquet2021stdnet} developed STDnet-ST for small object detection, achieving state-of-the-art performance but proposing the use of GANs for generating synthetic small objects.

\item Babiker et al \cite{Babiker2017AutomatedDH} developed an automated system for recognizing human activities using neural networks, achieving high recognition rates but needing testing in complex settings.

\item Deguerre et al \cite{deguerre2019fast} proposed a rapid object detection method in compressed videos, enhancing detection accuracy while reducing computational time but facing limitations with motion vector quality.

\item Ren et al \cite{ren2016faster} introduced the Faster R-CNN algorithm for fast and accurate object detection, achieving high performance but facing computational intensity issues.

\item Patil et al \cite{Patil2021AnEM} proposed a method for crowd analysis using motion patterns and SVMs, achieving high recognition rates but relying on controlled datasets.

\end{itemize}

These studies collectively advance the field of video surveillance, addressing various challenges and proposing innovative solutions, though they often highlight the need for further research to overcome limitations in real-world applications as some of them are presented in Table \ref{table1} .

\begin{table}[h!]
\caption{Common Limitations in Reviewed Surveillance Systems}
\centering
\begin{tabular}{|p{4cm}|p{8cm}|}
\hline
\textbf{Limitation} & \textbf{Description} \\
\hline
Fixed Camera Requirements & Systems like \cite{Abdelbaki2016} and \cite{suradkar2015automatic} work only with fixed cameras, limiting use in dynamic environments. \\
\hline
Lack of Real-time Processing & High computational complexity in systems like \cite{pal2023object} and \cite{sreenu2019intelligent} hinders real-time performance. \\
\hline
Indoor/Outdoor Constraints & Systems such as \cite{Babiker2017AutomatedDH} are designed for specific environments, limiting versatility. \\
\hline
Computational Requirements & High computational demands of algorithms like Faster R-CNN \cite{ren2016faster} and RetinaNet \cite{kapania2020multi} challenge real-time deployment on limited devices. \\
\hline
Generalization Issues & Limited datasets affect generalization in systems like \cite{boumediene2022detection}, \cite{aradhya2019object}, and \cite{lys2023development}. \\
\hline
\end{tabular}
\label{table1}
\end{table}

To address limitations in human detection and surveillance, several research directions are proposed, including hybrid approaches that combine background subtraction with deep learning or optical flow with Support Vector Machines (SVM) for improved accuracy in dynamic environments. Enhancing generalization through diverse training datasets, synthetic data, data augmentation, and transfer learning can also improve model performance in real-world conditions. Developing versatile models for both indoor and outdoor settings using mixed datasets and context-aware mechanisms can enhance surveillance system adaptability.

This paper focuses on optimizing storage in surveillance systems by recording only significant actions using advanced activity detection techniques. Unlike traditional systems like Hikvision\footnote{https://www.hikvision.com}, which use continuous or basic motion detection, our approach selectively captures important activities, thus reducing storage needs. We evaluate our method against Hikvision to demonstrate its effectiveness in reducing storage without sacrificing surveillance quality.

The study explores intelligent video recording optimization through various techniques, including motion detection, background subtraction, optical flow, and advanced object detection methods like YOLOv3 and Faster R-CNN. A hybrid approach that combines initial motion detection with precise object tracking is identified as optimal for performance and resource efficiency. Future research should refine these models, expand datasets, and utilize advanced training devices. While integrating these techniques for real-time surveillance remains challenging, our hybrid approach aims to achieve more efficient and effective surveillance.

\section{Methods}
The Hikvision Surveillance System (HikvisionSS) was selected as the baseline for comparison in this study due to its existing deployment within our institution, École supérieure en Sciences et Technologies de l'Informatique et du Numérique (ESTIN\footnote{https://estin.dz/}). The primary objective of this project is to optimize and reduce the storage space required for video recordings, which directly impacts the operational efficiency of the Hikvision system. As such, it was essential to evaluate the performance of our proposed approach in comparison to HikvisionSS, particularly in terms of the storage space required for recording, to ensure that our method provides tangible benefits within the context of its real-world application.
We have chosen the Hikvision surveillance system as the baseline for comparison due to its widespread use in our institution and its standard recording modes, which include continuous recording and motion detection-based recording. Our proposed method focuses on an intelligent approach that records only significant actions, identified through advanced activity detection algorithms. This strategic selection allows us to directly assess the improvements in storage efficiency offered by our approach.

\subsection{System Architecture}
\begin{figure}[h!]
    \centering    \includegraphics[width=\textwidth]{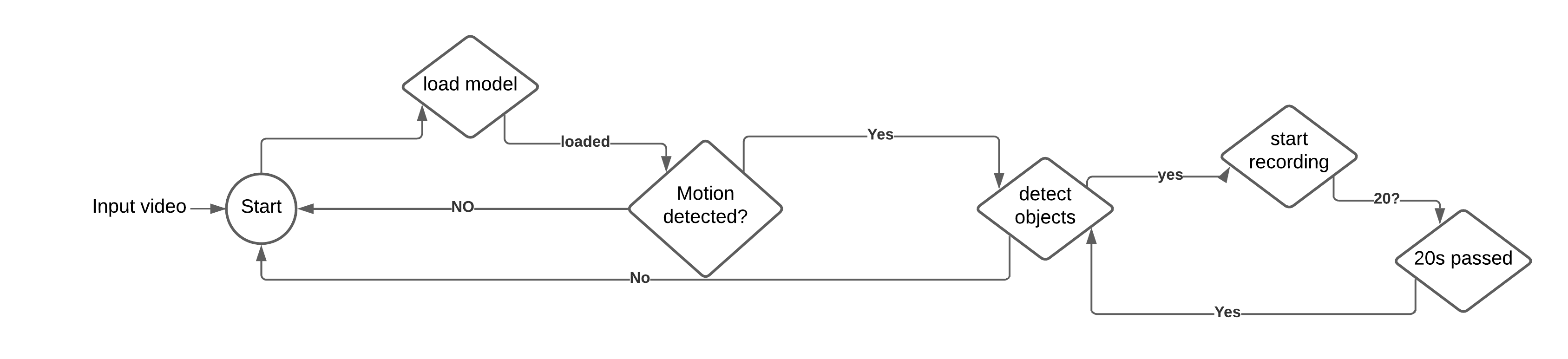}
    \caption{flowchart explaining the global architecture of the proposed system}
    \label{fig:fig1}
\end{figure}

The proposed system integrates motion detection and object detection in a hybrid approach. As illustrated in Figure \ref{fig:fig1}, the system begins with frame subtraction to identify regions of interest where motion is detected. These regions are then processed by the YOLO model, which detects and classifies objects, with a particular focus on human activity. This architecture ensures that only relevant scenes are recorded, significantly reducing unnecessary footage.

\begin{algorithm}
\caption{Object Detection Driven Surveillance Video Recording}
\label{alg:surveillance_recording}
\begin{algorithmic}
\REQUIRE Surveillance video stream
\ENSURE Recorded video files with object activity
\STATE Load the object detection Yolo model 
\STATE Initialize variables for motion detection and recording status
\WHILE{video stream exists}
    \STATE Read the current frame from the video stream
    \STATE Use the motion detection algorithm  to detect motion in the current frame
    \IF{ motion is detected}
       \STATE Use the object detection model to detect objects in the current frame
        \IF {an object is detected }
            \IF{not currently recording}
                \STATE Start recording and note the time of last detection and currently recording == True     
            \ENDIF
        \ENDIF
    \ENDIF
    \IF{currently recording}
        \IF{less than 20 seconds have passed since the last detection }
           \IF {objects are detected }
              \STATE Update the time of last detection   
            \ENDIF
            \STATE Write the current frame to the video file
        \ELSE
            \IF{no objects are detected }
                \STATE Stop recording and currently recording == False 
            \ELSE 
                \STATE Write the current frame to the video file and 
                Update the time of last detection   
            
            \ENDIF
        \ENDIF
    \ENDIF
\ENDWHILE
\IF{Stop recording button = True }
    \STATE Stop recording and currently recording == False 
\ENDIF
\end{algorithmic}
\end{algorithm}

\subsection{Activity Detection}
\subsubsection{Motion Detection:} Frame subtraction is employed to detect changes between consecutive frames, indicating potential activity. This method is computationally efficient and suitable for real-time applications \cite{manchanda2016analysis}.
To examine the different algorithms used for motion detection, we reviewed various works such as \cite{Abdelbaki2016} and \cite{Babiker2017AutomatedDH} utilizing background subtraction, \cite{Alajrami2019OnUA} employing optical flow, and \cite{pal2023object} proposing a composite block matching algorithm.

\subsubsection{Object Detection:} The YOLO model is used to detect and classify objects within the video frames. YOLOv9 is chosen for its balance between speed and accuracy, making it ideal for real-time surveillance.
\par An example of using the YOLO method is found in \cite{adarsh2020suspicious}, where the authors detect suspicious individuals and hostile behavior. They use the YOLO model, pre-trained on the COCO dataset \footnote{https://cocodataset.org/}, for human detection in video frames. These frames are then processed by ResNet-34 to recognize activities, achieving a precision of 82\%. Despite its effectiveness, the model's reliance on two deep learning models demands substantial computational resources

\par The workflow for recording surveillance videos driven by object detection is detailed in Algorithm \ref{alg:surveillance_recording}. The algorithm outlines how the system processes video streams, detects motion, and records activity only when objects of interest are present in the frame, ensuring efficient storage usage and effective monitoring.

\section{Implementation}
\subsection{Data Preparation and Analysis}

The data preparation involved collecting, labeling, and processing images for training the model. The dataset was sourced from various public repositories, focusing on human and car detection. Key data preparation steps included labeling objects using the Roboflow platform, preprocessing images (auto-orientation, resizing, contrast adjustment), and applying data augmentation techniques like grayscale conversion.

The dataset was split into training, validation, and testing sets using a 70:20:10 ratio. The dataset comprised 300 images for human detection from Kaggle and 100 images for car detection from a GitHub repository, supplemented by additional images sourced from Google Images and other websites. The preparation process involved auto-orienting the images, resizing them to 646x640 pixels, and applying contrast adjustment and grayscale conversion to a portion of the dataset. The final dataset included 2,664 images labeled as 'Person' and 1,735 as 'Car', ensuring a well-balanced distribution across object classes.

The analysis highlighted the efficiency of the data preparation process, ensuring that the training model was robust and well-balanced across different scenarios and object classes. Visual tools like heatmaps and histograms were used to further analyze the distribution of annotations and objects within the dataset.

\begin{figure}
\centering
\includegraphics[width=\linewidth]{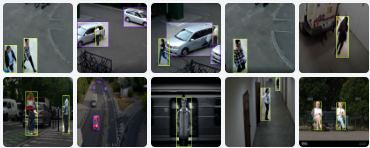}
\caption{Samples from the dataset used to train YOLO model}
\label{fig:fig2}
\end{figure}

Figure \ref{fig:fig2} presents samples from the dataset, showcasing the diversity of scenarios and object classes included.

\subsection{Model Training}
\subsubsection{Model Pre-training}
The YOLO algorithm was implemented by cloning its repository from GitHub. Pre-trained weights from the MS COCO dataset were used as a starting point for fine-tuning on the custom dataset. The pre-training process involved running a script with key parameters, including 8 CPU workers for faster data loading, GPU utilization, a batch size of 16, and 500 training epochs. The model was trained from scratch using specific configurations and hyperparameters, with mosaic augmentation disabled after 15 epochs to enhance focus on original images. This pre-trained model was then fine-tuned on the custom dataset.

\subsubsection{Model Fine-Tuning}
The fine-tuning of the YOLO model on the custom dataset was performed using a modified script, which involved resizing input images to 640x640 pixels, setting a batch size of 16, and training for 25 epochs. The data configuration file was pointed to the custom dataset, while the model configuration file specified the YOLO architecture. Pre-trained weights from the initial training on the MS COCO dataset were used as the starting point. 

\subsubsection{Evaluation of Training Results}
After completing the YOLO model training and fine-tuning, a thorough evaluation was performed using both qualitative and quantitative metrics to assess the model's performance. The validation process utilized a separate dataset to ensure the model's ability to generalize to unseen data. Key parameters for validation included using the prepared dataset, the best model weights from training, a batch size of 16, images resized to 640x640 pixels, a confidence threshold of 0.001, an IoU threshold of 0.5, and a maximum of 300 detections per image.
\subsection{Comparative Analysis of Trained Models} 
\par The comparative analysis of the trained models focuses on identifying the most suitable model for deployment by evaluating various performance metrics, such as precision, recall, and mean Average Precision (mAP), in addition to considering computational resources and deployment constraints. This analysis not only aids in selecting the optimal model but also informs iterative improvement strategies by highlighting areas for refinement. The 7th Model, with the highest precision (87.0\%), recall (82.0\%), and mAP (90.1\%), was selected for the proposed system, as it offers the best balance between precision and recall, ensuring accurate object detection and classification in video feeds.

\begin{table}[h]
\centering
\caption{Comparative Table of Trained Models}
\resizebox{\textwidth}{!}{%
\begin{tabular}{|c|c|c|c|c|c|c|}
\hline
\textbf{Model} & \textbf{Version} & \textbf{Dataset Size} & \textbf{Epochs} & \textbf{Precision} & \textbf{Recall} & \textbf{mAP (0.5)} \\
\hline
1st Model & 9  & 664 Images & 50 & 77.9\% & 73.7\% & 76.8\% \\
\hline
2nd Model & 8 & 664 Images & 50 & 70.0\% & 71.7\% & 74.3\% \\
\hline
3rd Model & 9  & 1177 Images & 30 & 81.6\% & 73.3\% & 80.9\% \\
\hline
4th Model & 9 & 1469 Images & 30 & 85.3\% & 75.2\% & 82.4\% \\
\hline
5th Model & 9  & 1294 Images & 30 & 83.6\% & 75.1\% & 82.2\% \\
\hline
6th Model & 9 & 1622 Images & 25 & 86.5\% & 76.7\% & 85.5\% \\
\hline
\textbf{7th Model} & \textbf{9} & \textbf{1920 Images} & \textbf{25} & \textbf{87.0\%} & \textbf{82.0\%} & \textbf{90.1\%} \\
\hline
8th Model & 9 & 1927 Images & 100 & 85.1\% & 82.6\% & 88.7\% \\
\hline
\end{tabular}
}
\label{tab:comparison}
\end{table}

\subsection{Model Deployment}
\par The deployment process involves preparing the chosen YOLOv9 based model by converting it into a deployable format and configuring it for integration. Once deployed, the model is accessible via an API, allowing seamless integration into the application environment. This setup ensures the model performs accurate inference tasks and provides reliable object detection capabilities, making it ready for real-world application scenarios.

\subsection{Model Evaluation and Training Visualization}

The model's performance was evaluated using key metrics such as Precision (P), Recall (R), mean Average Precision at IoU threshold 0.5 (mAP50), and mean Average Precision across IoU thresholds from 0.5 to 0.95 (mAP50-95). Precision, which indicates the accuracy of positive predictions, was 0.869 overall (Car: 0.855, Person: 0.884). Recall, reflecting the model's ability to detect all relevant instances, was 0.824 overall (Car: 0.83, Person: 0.819). The mAP50 was 0.891 (Car: 0.899, Person: 0.883), showing high precision and recall at an IoU threshold of 0.5, while the mAP50-95 was 0.558 (Car: 0.638, Person: 0.478), demonstrating robustness across different IoU settings. These metrics, detailed in Table \ref{tab:model_performance_metrics}, highlight the model’s strong performance in detecting cars and persons with high precision and recall, and reasonable generalization across IoU thresholds.

\begin{table}[h]
    \centering
        \caption{Model Performance Metrics for Precision, Recall, and mAP.}
    \begin{tabular}{|c|c|c|c|c|}
        \hline
        \textbf{Category} & \textbf{Precision} & \textbf{Recall} & \textbf{mAP50} & \textbf{mAP50-95} \\ \hline
        Car    & 0.855 & 0.83  & 0.899 & 0.638 \\ \hline
        Person & 0.884 & 0.819 & 0.883 & 0.478 \\ \hline
        Overall & 0.869 & 0.824 & 0.891 & 0.558 \\ \hline
    \end{tabular}   \label{tab:model_performance_metrics}
\end{table}

The confusion matrix, which evaluates the model's performance by comparing predicted labels to true labels, includes metrics such as True Positives (TP), False Positives (FP), False Negatives (FN), and True Negatives (TN). The model demonstrates efficient performance, with a pre-processing time of 0.5 milliseconds, an inference time of 25.1 milliseconds, and Non-Maximum Suppression (NMS) time of 5.2 milliseconds per image. The inference time is particularly significant, as it reflects the model's capability for real-time image processing, which is crucial for applications requiring rapid decisions.

\begin{table}[h]
    \caption{Performance Speed Metrics.}
    \centering
    \begin{tabular}{|c|c|}
        \hline
        \textbf{Performance Metric} & \textbf{Time (milliseconds per image)} \\ \hline
        Pre-processing Time         & 0.5   \\ \hline
        Inference Time              & 25.1  \\ \hline
        Non-Maximum Suppression (NMS) Time & 5.2   \\ \hline
    \end{tabular}
    \label{tab:performance_speed}
\end{table}

\begin{figure}[h!]
    \centering
    \includegraphics[width=1\textwidth]{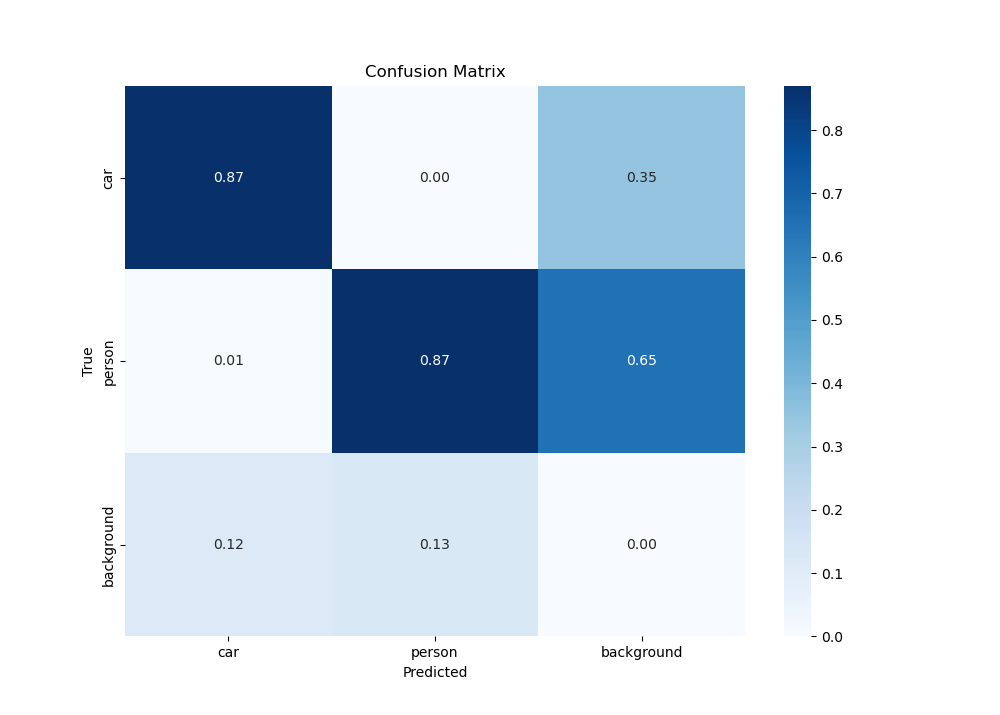}
    \caption{Confusion matrix for the proposed video recording optimization method applied to surveillance footage}.
    \label{fig:3}
\end{figure}

Figure \ref{fig:3} shows the confusion matrix for activity classification (car, person, and background) in surveillance footage. The diagonal elements represent correctly classified instances, while the off-diagonal elements correspond to misclassifications. The matrix indicates an accuracy of 87\% for detecting cars and persons, with slightly lower accuracy for the background class. Misclassifications mainly occurred between 'person' and 'background,' highlighting areas for potential model improvement.

The training progress is illustrated through plots showing the evolution of loss components and performance metrics over epochs. The bounding box regression loss (train/box\_loss) decreased from 1.5 to 1.1, indicating improved accuracy in predicting bounding box coordinates. The classification loss (train/cls\_loss) fell from 1.6 to about 0.6, reflecting enhanced classification performance. The Distribution Focal Loss (train/dfl\_loss) also decreased from 1.5 to 1.25, demonstrating increased precision. Precision (metrics/precision) and recall (metrics/recall) metrics improved from 0.65 to approximately 0.85 and 0.82, respectively. Validation losses (val/box\_loss, val/cls\_loss, val/dfl\_loss) followed similar trends, indicating good generalization. The mean Average Precision at an IoU threshold of 0.5 (metrics/mAP\_0.5) increased from 0.45 to 0.90, and the mean Average Precision across IoU thresholds from 0.5 to 0.95 (metrics/mAP\_0.5:0.95) rose from 0.40 to 0.75, highlighting the model’s robust performance.

\begin{figure}[h!]
    \centering
    \includegraphics[width=\textwidth]{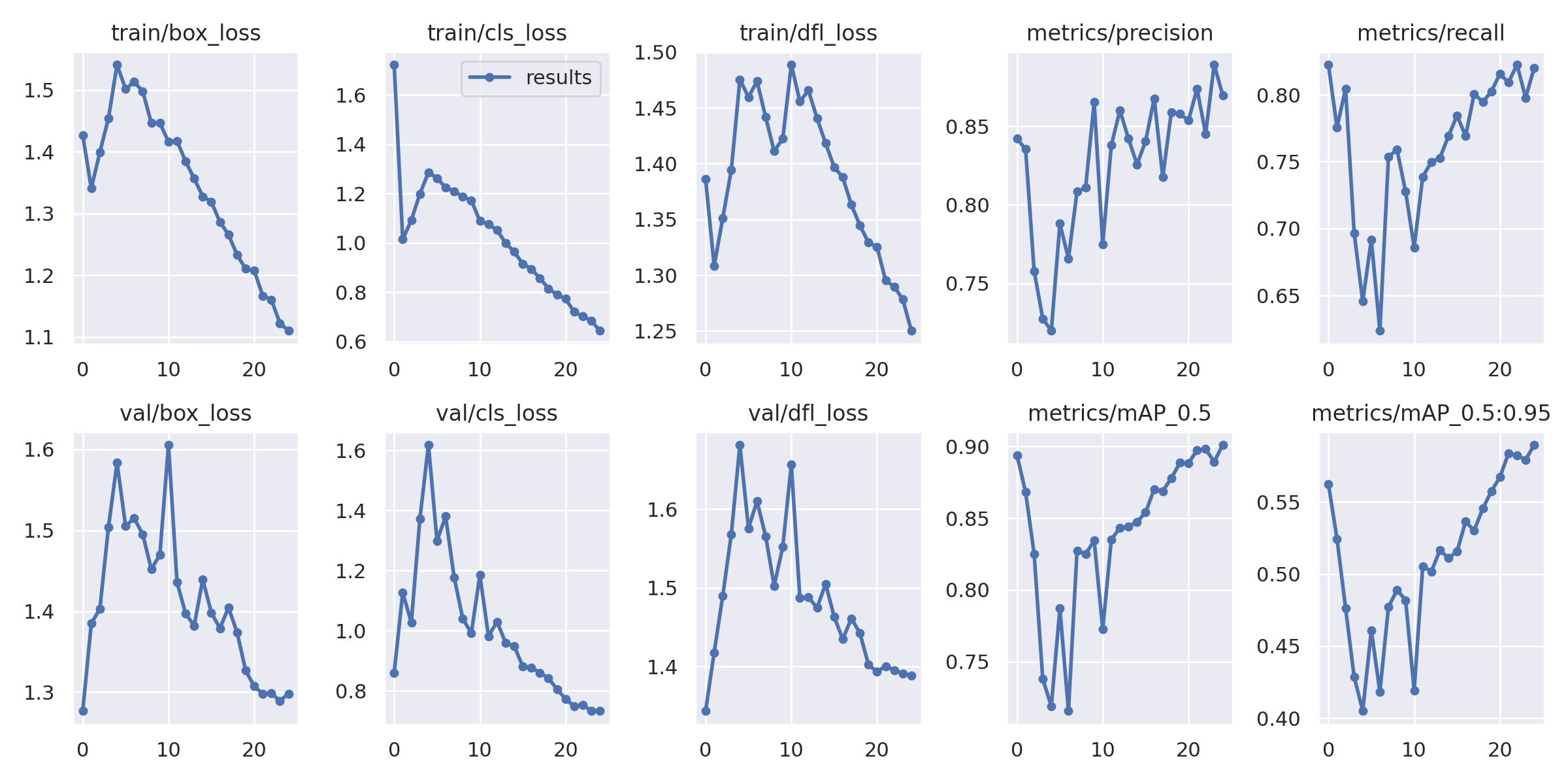}
    \caption{Training and Validation Performance Metrics of the 7th YOLOv9 based Model.}
    \label{fig:4}
\end{figure}

Figure \ref{fig:4} displays training and validation losses, precision, recall, and mean Average Precision (mAP) metrics over 25 epochs. The training and validation losses decrease consistently, indicating effective learning. Precision and recall improve steadily, with recall nearing 0.90 by the end of training. The mAP metrics also show positive trends, with mAP\_0.5 surpassing 0.8. These results suggest the model continues to improve, with further training potentially enhancing performance. The visualizations in Figure \ref{fig:4} provide insights into the model’s learning progress, crucial for assessing overfitting or underfitting. Overall, the best YOLOv9 model, fine-tuned with the custom dataset, exhibits high accuracy and efficiency, making it suitable for deployment in the proposed intelligent video recording optimization solution.

\section{Deployment and Results}

\subsection{Storage Optimization}.
This setup involves two different scenarios: one using the proposed intelligent recording approach and the other employing the Hikvision surveillance system's traditional continuous recording and motion detection features. The primary objective is to quantify the reduction in storage space achieved by our method, without sacrificing the accuracy and reliability of the surveillance footage.
To validate the performance of the proposed application and the study objective, which focuses on the Optimization of Recording Storage for Surveillance Systems, a comparative study was conducted using Hikvision monitoring software\footnote{https://www.hikvision.com/en/support/download/software/} at ESTIN. Both systems were tested simultaneously under identical conditions to measure their effectiveness in optimizing storage by comparing the length of recorded videos. All recordings were made using a unified format (MP4, codec H.264, 1280x720, 16:9, 30 fps, bitrate 7 703 kbps).

\subsubsection{Equipment and Configuration}
Two types of cameras were used: a fixed camera and a flexible dome camera, positioned strategically in front of ESTIN's main gate and the building of Labs' building. The Hikvision system was configured with its specific hardware and software setup, while the proposed system was deployed on a local machine, specifically an i7 Core ThinkPad. This setup created a controlled environment to accurately measure and compare the performance of both systems in terms of storage optimization.

For the experimental evaluation, two distinct video records were used. The first video was a real-time test conducted live with the first camera
This period of time in a workday, particularly from 2 PM to 3 PM, is likely one of the most active for students, teachers, and workers at ESTIN. The first video captures the scene in front of the main gate of ESTIN during this busy period, from 2 PM to 3 PM. The second video was recorded offline, after the real-time scene was captured. This second recording was done using a flexible camera positioned in front of the Labs' building from 2:30 PM to 2:45 PM. This dual approach allowed us to test the proposed method under both live and post-recording conditions, providing a comprehensive evaluation of its performance.

\subsubsection{Real-Time Test}
\begin{itemize}
    \item Objective: Compare the storage optimization between the proposed system and the Hikvision software over a one-hour period of live surveillance with motion detection enabled.

    \item Methodology: Both systems were launched simultaneously and ran for one hour with motion detection enabled on both.

    \item Metrics Collected: Length of the recorded videos from both systems.
\end{itemize}

\begin{figure}[h!]
  \centering  \includegraphics[trim=0 50 200 0, clip, scale=0.5]{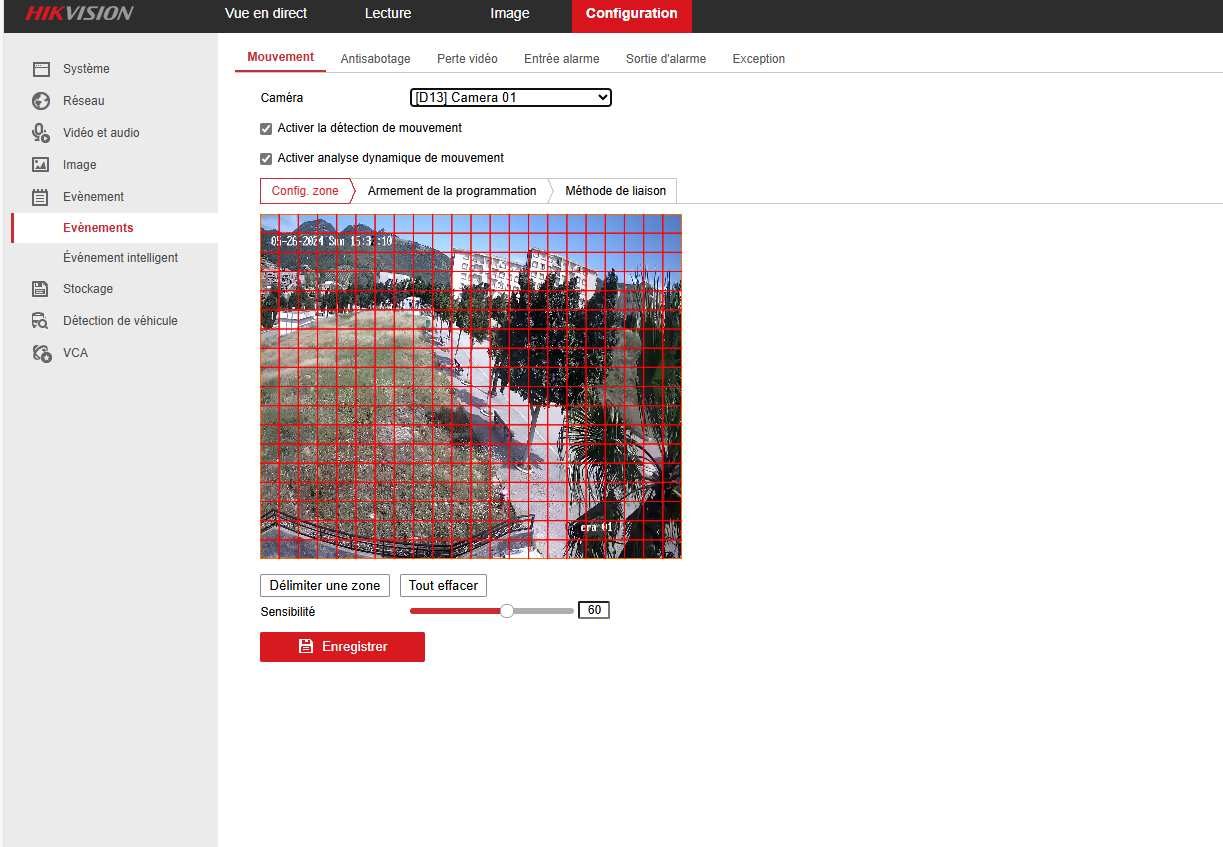}
  \caption{Screenshot of the Hikvision System with Motion Detection Enabled}
  \label{fig:enter-label}
\end{figure}

\subsubsection{Video Upload Test}
\begin{itemize}

    \item Objective: Compare the performance of the proposed system and the Hikvision software using a pre-recorded video.

    \item Methodology: A 15-minute video was processed by both systems with motion detection enabled.

    \item Metrics Collected: Length of the resulting videos after processing.
\end{itemize}

Due to practical constraints, the number of video records used in this study was limited. However, the selected videos were chosen for their representativeness in assessing the effectiveness of the proposed system in optimizing storage space compared to the HikvisionSS baseline. Despite these limitations, the focus of our analysis remains on demonstrating the efficiency and applicability of our approach within the context of real-world surveillance scenarios.

\subsubsection{Analysis of Storage Efficiency}
The performance of the two systems was evaluated based on the following criteria:

\begin{itemize}

    \item Recorded Video Length: The total duration of the videos recorded by each system under the same conditions. A shorter recorded length indicates better storage optimization.
    \item Detection Accuracy: The ability to correctly identify and record relevant activities (human and vehicular) without missing any significant events.
\end{itemize}

\begin{table}[h!]
\centering
\caption{Comparison of Recorded Lengths and Storage for Different Test Types}
\begin{tabular}{|>{\centering\arraybackslash}m{5.3cm}|>{\centering\arraybackslash}m{2cm}|>{\centering\arraybackslash}m{1.7cm}|>{\centering\arraybackslash}m{2cm}|}
\hline
\textbf{Test Type} & \textbf{System} & \textbf{Video Length (minutes \& seconds)} & \textbf{Storage Size (MegaBytes)} \\
\hline
\multirow{2}{5.3cm} {1-Hour Real-Time Test (live)} & Hikvision SS & 60m \& 00s & 1 917\\
\cline{2-4}
& Proposed & 13m \& 45s & 769\\
\hline
\multirow{2}{5.3cm} {15-Minute Video Test (recorded)} & Hikvision SS & 6m \& 53s & 379\\
\cline{2-4}
& Proposed & 6m \& 20s & 355\\
\hline
\end{tabular}
\label{table2}
\end{table}

To evaluate the efficiency of the proposed method, a comparative analysis was conducted using two test types: a 1-hour real-time test and a 15-minute video test. The results in Table \ref{table2} compare the recorded lengths and storage sizes between the traditional Hikvision surveillance system and the proposed activity detection system. In the 1-hour real-time test, the Hikvision system recorded 60 minutes, consuming 1 917 MB, while the proposed system recorded only 13 minutes and 45 seconds, using 769 MB. This discrepancy is largely due to Hikvision's motion detection being highly sensitive to minimal motion, such as the movement of trees and palms caused by the wind, which triggered recording frequently. In contrast, the proposed system only initiated recording after detecting significant motion, using object detection to ensure that the recorded scenes contained relevant objects like cars or persons. In the second 15-minute video test, there was no wind, and both systems recorded only significant motions when detected. Therefore, the Hikvision system recorded 6 minutes and 53 seconds, consuming 379 MB, whereas the proposed system recorded 6 minutes and 20 seconds, using 355 MB.

These results demonstrate significant storage savings by recording only scenes with detected activity, reducing storage requirements without compromising important data. The storage savings chart underscores the efficiency of the proposed method compared to traditional continuous recording.

\begin{figure}[h!]
\centering
\includegraphics[width=0.4\textwidth]{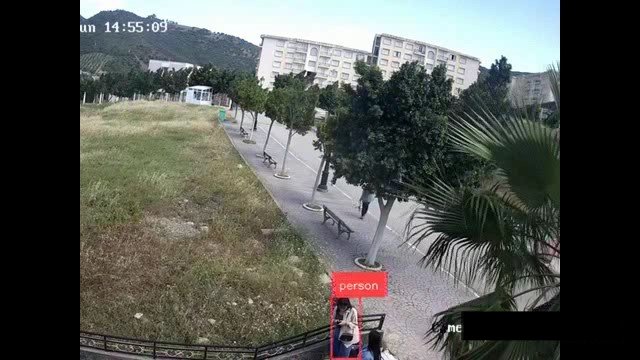}
\includegraphics[width=0.46\textwidth]{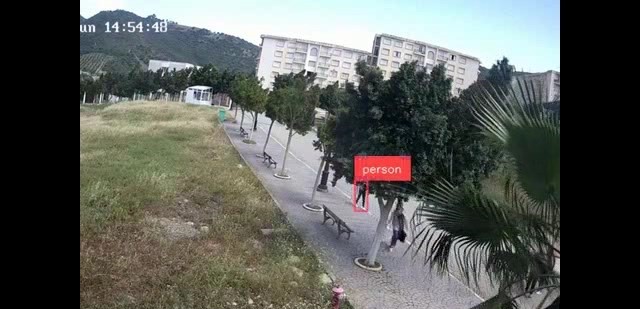}
\includegraphics[width=0.4\textwidth]{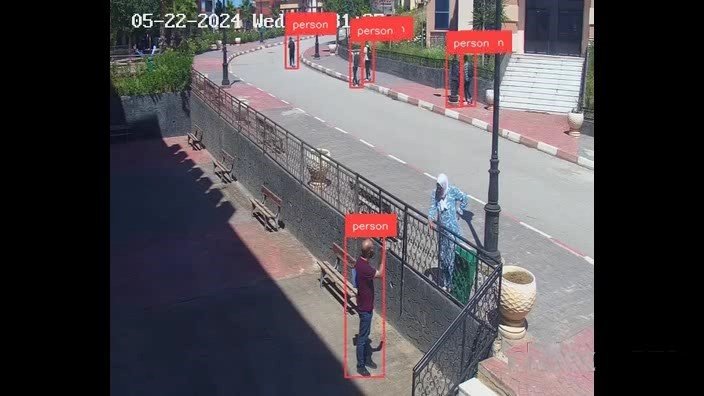}
\includegraphics[width=0.46\textwidth]{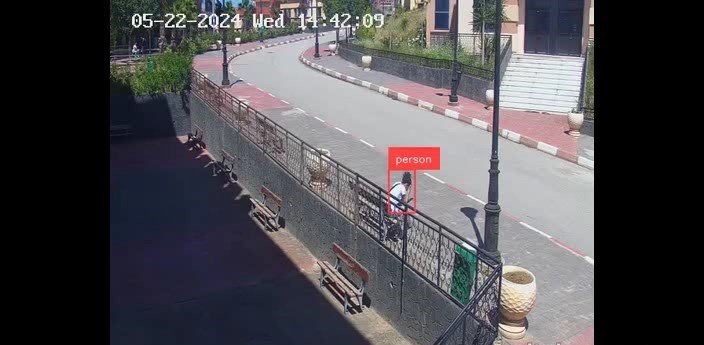}
\caption{Samples of real-time validation}
\label{fig:fig6}
\end{figure}

However, the proposed system has some limitations, such as false positives and false negatives in adverse weather conditions like strong winds, which can cause motion artifacts that affect the accuracy of detection as it is illustrated in Figure \ref{fig:3} and in some amples of Figure \ref{fig:fig6}.

The validation results demonstrate that the proposed system significantly outperforms the Hikvision software in optimizing storage. In the one-hour real-time test, the proposed system recorded only 13 minutes and 45 seconds of video, compared to Hikvision's 60 minutes, even with Hikvision's intelligent motion detection option enabled. This highlights the proposed system's superior ability to filter out irrelevant footage and focus on significant activities.

In a 15-minute video upload test, the proposed system recorded 6 minutes and 20 seconds, while Hikvision recorded 6 minutes and 53 seconds. Although Hikvision captured the necessary scenes, it encountered some bugs. The proposed system proved more efficient by accurately detecting and recording pertinent activities, thereby minimizing storage usage.

These results confirm the effectiveness of the proposed model in optimizing recording storage for surveillance systems. Compared to Hikvision, which relies on continuous or basic motion-based recording, our method significantly reduces the storage space required by recording only meaningful activities. This optimization does not compromise the ability to capture critical events, as evidenced by the comparative analysis, making our approach a valuable tool for enhancing the efficiency and cost-effectiveness of surveillance operations.

\section{Conclusion}
In conclusion, this work effectively demonstrates that intelligent activity detection can optimize storage space in surveillance systems. By selectively recording only meaningful actions, our approach significantly reduces storage requirements compared to traditional systems like Hikvision, which rely on continuous or motion-based recording.

This study developed a hybrid system that combines motion detection via successive frame subtraction with the YOLOv9 model for detecting significant activities. YOLOv9 was selected for its superior detection capabilities and efficiency, achieving an accuracy of 87\%. The combination of motion detection for initial screening with advanced activity detection optimizes both performance and resource usage, particularly benefiting the recording system at ESTIN by addressing storage and resource management challenges. Validation results showed substantial optimization in both time and storage space.

Future work should focus on refining model structures to further enhance detection precision and efficiency, leveraging more powerful training devices and larger, high-quality datasets. While our project made significant strides, further advancements are possible with additional resources and improved data. The methodologies developed hold potential applications beyond security, benefiting any field requiring real-time object detection and intelligent video recording.

Although the scope of our experiments was limited due to practical constraints, the results highlight the potential of the proposed method for optimizing video recording in surveillance systems. Future studies should consider expanding the dataset and exploring additional comparative methods to further validate the system's performance.

\section*{Acknowledgments}
We express our deepest gratitude to Mr. Nabil Aoughlis, Ms. Lylia Aberkane and Mr. Ridha Chekroune for their guidance and support.

\bibliographystyle{IEEEtran}
\bibliography{mybibfile}

\end{document}